\documentclass{article}
\usepackage{spconf,amsmath,graphicx,hyperref}
\usepackage{booktabs}
\usepackage{tabularx}
\usepackage{caption}
\usepackage{array} 
\usepackage{colortbl}
\usepackage{xcolor}
\usepackage{makecell} 
\usepackage{amsmath,amssymb}
\usepackage{algpseudocode}
\usepackage{algorithm}
\usepackage[T1]{fontenc}
\usepackage[utf8]{inputenc}

\newcolumntype{Y}{>{\centering\arraybackslash}X}
\definecolor{bestcolor}{RGB}{255,200,120}

\DeclareUnicodeCharacter{0097}{---} 

\title{PTTA: A Pure Text-to-Animation Framework for High-Quality Creation}
\name{Ruiqi Chen \qquad Kaitong Cai \qquad Yijia Fan \qquad Keze Wang}
\address{Sun Yat-sen University}

\begin{document}
\maketitle
\begin{abstract}

Traditional animation production involves complex processes and significant manual labor costs. While existing video generation models such as Sora, Kling, and CogVideoX excel at generating natural videos, they face limitations when applied to animation generation tasks. Recent efforts, like Anisora, have achieved promising results by fine-tuning image-to-video (I2V) models for animation styles. However, analogous work in the text-to-video (T2V) domain remains largely unexplored. In this paper, we first construct a small-scale, high-quality animation text-video paired dataset. Then, based on the pretrained T2V model HunyuanVideo, we perform fine-tuning to obtain an animation-style T2V model. Extensive visual evaluations across multiple dimensions demonstrate that our fine-tuned model outperforms comparable approaches in generating animation videos.
\end{abstract}
\begin{keywords}
Multimodal, Text-to-Video Generation, Fine-tuning, Anime style
\end{keywords}
\section{Introduction}
\label{sec:intro}

In recent years, generative AI has achieved revolutionary breakthroughs in video creation, with large-scale models like Sora and Kling capable of producing highly realistic natural scenes. However, the domain of animation—a unique and highly expressive art form—presents a notable challenge for these advancements. The essence of animation lies in its non-realistic physics, exaggerated character expressions, and highly stylized aesthetics—qualities that general-purpose models trained on real-world data struggle to capture. Meanwhile, traditional animation production remains heavily reliant on labor-intensive manual drawing, leading to high costs and long production cycles. This establishes an urgent need for automated, high-efficiency animation generation technologies.

To bridge this gap, researchers have begun adapting existing models for animation. However, these efforts, such as AniSora\cite{jiang2024anisora,z10}, face a critical bottleneck: they often require an image as an additional condition (Image-to-Video) to ensure the coherence and plausibility of the generated content. This reliance on visual inputs not only constrains creative freedom but also falls short of the ultimate goal of generating animation from scratch purely from textual descriptions. \textbf{The core challenge, therefore, is to develop a model that can understand the unique language of animation and generate high-quality, dynamic content purely from text prompts (Text-to-Video).}

To address this challenge, we introduce \textbf{PTTA (Pure Text-to-Animation)}, a framework specifically designed for high-quality, purely text-driven animation generation. Our work fundamentally tackles two primary obstacles: data scarcity and model adaptation. First, we constructed a new dataset of over 12,000 high-quality text-animation video pairs, providing a solid foundation for model training. Subsequently, we leverage the powerful open-source model HunyuanVideo\cite{z4} and employ an advanced asymmetric fine-tuning strategy, \textbf{HydraLoRA}\cite{tian2024hydralora,Z5,z8,z12}, for parameter-efficient adaptation. This strategy enables the model to effectively learn diverse animation styles and achieve strong generalization even with a modestly-sized dataset. The ultimate goal of PTTA is to eliminate the dependency on image inputs, enabling true ``what-you-see-is-what-you-get'' animation creation from text alone.

Our work is designed to address three core challenges that have hindered progress in text-to-animation generation. First, the scarcity of high-quality, large-scale text-video datasets for animation presents a fundamental roadblock for research. Second, the immense stylistic diversity of animation makes it difficult for models to generalize from limited data; efficiently adapting large pre-trained models to this niche domain is a significant hurdle. Finally, existing animation-focused methods often rely on image conditioning as a crutch, failing to achieve true creative control from text alone. Our contributions are direct solutions to these challenges:

\begin{itemize}
    \item \textbf{A High-Quality Animation Dataset.} We construct and release a dataset of over 12,000 curated text-animation video pairs to address the critical data scarcity in this domain.
    \item \textbf{An Efficient Domain Adaptation Paradigm.} We validate the effectiveness of the HydraLoRA fine-tuning strategy for animation T2V, establishing an efficient paradigm for adapting large models to stylistically diverse domains.
    \item \textbf{A Pure Text-to-Animation Model.} We present PTTA, a model capable of generating high-quality animated videos from text alone without any visual conditioning, achieving state-of-the-art performance.
\end{itemize}

\section{Related work}
\label{sec:relatedwork}
\begin{figure*}[t]
  \centering
\includegraphics[width=0.85\linewidth]{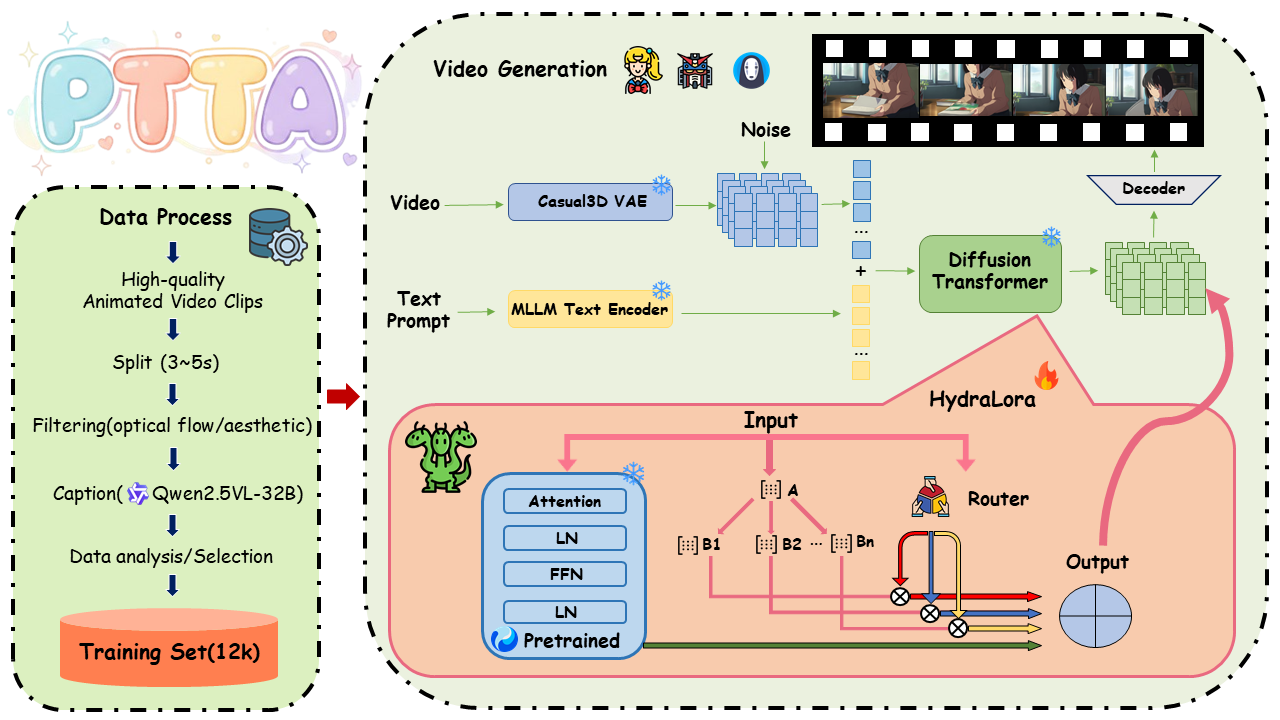}
  \caption{Overview: We present PTTA, a pure text-conditioned animation video generation model built upon a self-constructed dataset of over 12,000 high-quality text-video pairs. The data processing pipeline is illustrated on the left of the figure. The model is efficiently fine-tuned on HunyuanVideo using the HydraLoRA strategy, as depicted by the multi-headed Hydra icon, employing an asymmetric multi-branch LoRA framework to effectively enhance both the model’s generalization and generation capabilities.}
  \label{fig:img3}
\end{figure*}
\textbf{Text-to-Video Models.}
Text-to-video (T2V) generation has historically lagged behind its text-to-image (T2I) counterpart due to video data scarcity and complexity. 
Recent advancements in T2I, driven by diffusion and other generative models\cite{rombach2022high,CF-VLM,z13}, have provided a blueprint for T2V. Researchers have successfully adapted T2I architectures by integrating temporal modules to capture motion, enabling high-quality video generation\cite{blattmann2023align}.

Current T2V architectures, such as cascaded structures\cite{singer2022make} and latent diffusion models\cite{z7,diffu,z6}, often rely on joint image-video training strategies\cite{ho2022video} and build upon pretrained T2I models like Stable Diffusion\cite{rombach2022high}, DALL·E2\cite{ramesh2022hierarchical}.
While effective for general video synthesis, these models struggle with the unique stylistic demands of animation. Moreover, although recent advances in agent-driven coordination highlight the importance of dynamic adaptation in complex tasks\cite{KABB,gam}, existing animation-focused models like AniSora often require an image condition, limiting creative freedom. Our work, PTTA, addresses this gap by proposing a purely text-conditioned T2V model tailored for animation.

\textbf{Animation Video Datasets.}
High-quality text-video pairs are essential for T2V , yet they are particularly scarce in the animation domain. 
Early datasets like ATD-12K\cite{siyao2021deep} and AVC\cite{wu2022animesr} are small-scale and were primarily designed for tasks such as interpolation, not generative modeling. 
Even the large-scale Sakuga42M dataset\cite{pan2024sakuga}, with 1.2 million clips, is insufficient for pretraining high-quality models due to prevalent low-resolution content and short durations (under 2 seconds for 80\% of clips). 
To overcome these limitations, we constructed a new, high-quality dataset of over 12,000 text-video pairs specifically for fine-tuning a large pretrained model like HunyuanVideo, ensuring both feasibility and performance.

\section{Methodology}
\label{sec:methodology}

High-quality text-video pairs form the foundation of video generation models, as demonstrated in recent studies. In this section, we detail the construction of our animation training dataset and the underlying methodology for building our purely text-driven animation generation model.

\subsection{Dataset Construction and Curation}
\label{ssec:dataset}

Our process begins with data acquisition and curation. As illustrated in Figure~\ref{fig:img3}, we collected a large-scale set of high-resolution animation videos and segmented them into 3-5 second clips. These clips were then subjected to a rigorous filtering process based on two key metrics: optical flow scores and aesthetic scores. To formalize this, we define a selection score \(S\) for each clip:
\begin{equation}
S = \alpha \cdot S_{\text{optical}} + (1 - \alpha) \cdot S_{\text{aesthetic}}, \quad \text{clip is selected if } S \geq \theta,
\label{eq:filter_score}
\end{equation}
where \( S_{\text{optical}} \) and \( S_{\text{aesthetic}} \) represent the optical flow and aesthetic scores, respectively. The weighting factor \(\alpha\) balances their importance, and \(\theta\) is a predefined inclusion threshold. In our case, we set \( \alpha\) = 0.7 and \( \theta\) = 10 to ensure a high-quality selection of clips. Following this filtering, we employed the Qwen2.5-VL-72B model to generate descriptive text captions. This pipeline yielded a high-quality dataset of over 12,000 text-video pairs, providing a robust foundation for model training.

\subsection{Core Architecture and Theoretical Formulation}
\label{ssec:architecture}

Our work builds upon HunyuanVideo, a latent diffusion model (LDM) with over 13 billion parameters. The fundamental principle of an LDM is to train a denoising network in a compressed latent space. This process involves a forward noising process that gradually adds Gaussian noise to the data, and a reverse denoising process where the model learns to predict and remove this noise.

The simplified training objective for the DiT-based backbone, conditioned on text embeddings \(\mathbf{c}\), is to minimize the mean squared error between the true noise \(\boldsymbol{\epsilon}\) and the predicted noise \(\boldsymbol{\epsilon}_{\theta}\):
\begin{equation}
\small
\mathcal{L}_{\text{LDM}} = \mathbb{E}_{\mathbf{z}_0, \mathbf{c}, \boldsymbol{\epsilon}, t} \left[ \left\| \boldsymbol{\epsilon} - \boldsymbol{\epsilon}_{\theta}(\mathbf{z}_t, t, \mathbf{c}) \right\|^2 \right],
\label{eq:diffusion_loss}
\end{equation}
where \(\mathbf{z}_0\) is the latent encoding of the initial video provided by the VAE, \(\mathbf{z}_t = \sqrt{\bar{\alpha}_t}\mathbf{z}_0 + \sqrt{1-\bar{\alpha}_t}\boldsymbol{\epsilon}\) is the noisy latent at timestep \(t\), and \(\boldsymbol{\epsilon}_{\theta}\) is the denoising network parameterized by \(\theta\). The model's architecture, which executes this objective, consists of three main components:
\textbf{Causal 3D Variational Autoencoder (VAE)} compresses high-resolution videos into a compact latent space \(\mathbf{z}\), significantly reducing computational load while preserving spatiotemporal information.
\textbf{Text Encoder}, a multimodal large language model (MLLM), transforms text prompts into rich conditioning embeddings \(\mathbf{c}\) that effectively guide the generation process.
\textbf{DiT-based Backbone} (\(\boldsymbol{\epsilon}_{\theta}\)) operates on the noisy latents \(\mathbf{z}_t\) and text embeddings \(\mathbf{c}\) to predict the noise \(\boldsymbol{\epsilon}\), as formulated in Eq.~(\ref{eq:diffusion_loss}).

\subsection{Parameter-Efficient Fine-Tuning with HydraLoRA}
\label{ssec:hydralora}

To adapt the pretrained HunyuanVideo model to the animation domain efficiently, we employ HydraLoRA, a parameter-efficient fine-tuning (PEFT) technique. Standard LoRA modifies a pretrained weight matrix \(\mathbf{W}_0\) by adding a low-rank decomposition:
\begin{equation}
\mathbf{h}' = (\mathbf{W}_0 + \Delta\mathbf{W})\mathbf{x} = (\mathbf{W}_0 + \mathbf{B}\mathbf{A})\mathbf{x},
\label{eq:lora}
\end{equation}
where \(\mathbf{A} \in \mathbb{R}^{r \times k}\) and \(\mathbf{B} \in \mathbb{R}^{d \times r}\) are the trainable low-rank matrices, with rank \(r \ll \min(d, k)\).
HydraLoRA extends this by decomposing the update \(\Delta\mathbf{W}\) into multiple heads with a shared \(\mathbf{A}\) matrix, which is more effective for learning heterogeneous domain knowledge:
\begin{equation}
\Delta\mathbf{W} = \left( \sum_{i=1}^{N} \omega_i \cdot \mathbf{B}_i \right) \mathbf{A},
\label{eq:hydralora}
\end{equation}
where \(N\) is the number of specialist heads (e.g., for different animation styles or elements), and \(\omega_i\) are learnable or fixed weights. This structure allows the shared matrix \(\mathbf{A}\) to capture general animation features, while individual \(\mathbf{B}_i\) matrices specialize in distinct stylistic variations.

During fine-tuning, we freeze the original model weights \(\mathbf{W}_0\) and only update the parameters of the HydraLoRA adapters, \(\theta_{\text{Hydra}} = \{\mathbf{A}, \mathbf{B}_1, \dots, \mathbf{B}_N\}\). The optimization objective is to minimize the LDM loss on our animation dataset:
\begin{equation}
\small
\min_{\theta_{\text{Hydra}}} \mathcal{L}_{\text{LDM}}(\theta_0 + \Delta\theta(\theta_{\text{Hydra}})),
\label{eq:hydra_update}
\end{equation}
where \(\theta_0\) represents the frozen pretrained parameters and \(\Delta\theta(\theta_{\text{Hydra}})\) is the update from HydraLoRA. This approach enables PTTA to acquire specialized knowledge of the animation domain while retaining the powerful generative capabilities of the base model. The overall training procedure is outlined in Algorithm~\ref{alg:training}.
\begin{algorithm}[t]
\caption{PTTA Fine-tuning with HydraLoRA}
\label{alg:training}
\begin{algorithmic}[1]
\Require Pretrained parameters $\theta_0$, Dataset $\mathcal{D}$, Num heads $N$
\State Initialize HydraLoRA adapters $\theta_{\text{Hydra}}$; Freeze $\theta_0$
\For{each training epoch}
    \For{each batch $(\mathbf{v}, \mathbf{y})$ in $\mathcal{D}$}
        \State $\mathbf{z}_0, \mathbf{c} \gets \text{VAE.encode}(\mathbf{v}), \text{TextEncoder.encode}(\mathbf{y})$
        \State $t \sim \text{Uniform}(1, T)$, $\boldsymbol{\epsilon} \sim \mathcal{N}(0, \mathbf{I})$
        \State $\mathbf{z}_t \gets \sqrt{\bar{\alpha}_t}\mathbf{z}_0 + \sqrt{1-\bar{\alpha}_t}\boldsymbol{\epsilon}$
        \State $\boldsymbol{\epsilon}_{\theta} \gets \text{DiT}(\mathbf{z}_t, t, \mathbf{c}; \theta_0 + \Delta\theta(\theta_{\text{Hydra}}))$
        \State $\mathcal{L} \gets \| \boldsymbol{\epsilon} - \boldsymbol{\epsilon}_{\theta} \|^2$
        \State Update $\theta_{\text{Hydra}}$ using $\nabla_{\theta_{\text{Hydra}}} \mathcal{L}$
    \EndFor
\EndFor
\end{algorithmic}
\end{algorithm}
\setlength{\tabcolsep}{3pt}
\begin{figure}[t]
\centering
\setlength{\tabcolsep}{0pt}
\renewcommand{\arraystretch}{0.5}
\begin{tabular}{c c c c c c}
\raisebox{0.2\height}{\rotatebox{90}{\fontsize{6pt}{6pt}\selectfont HYBase}} &
\includegraphics[width=0.19\linewidth]
{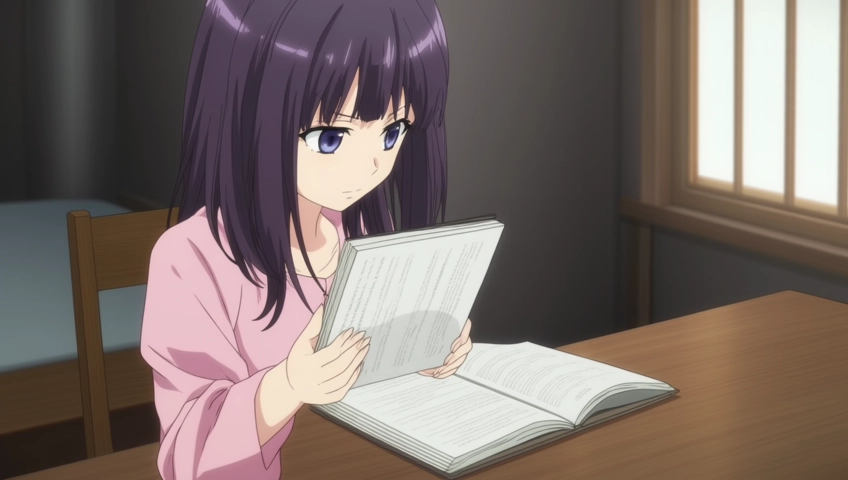} &
\includegraphics[width=0.19\linewidth]{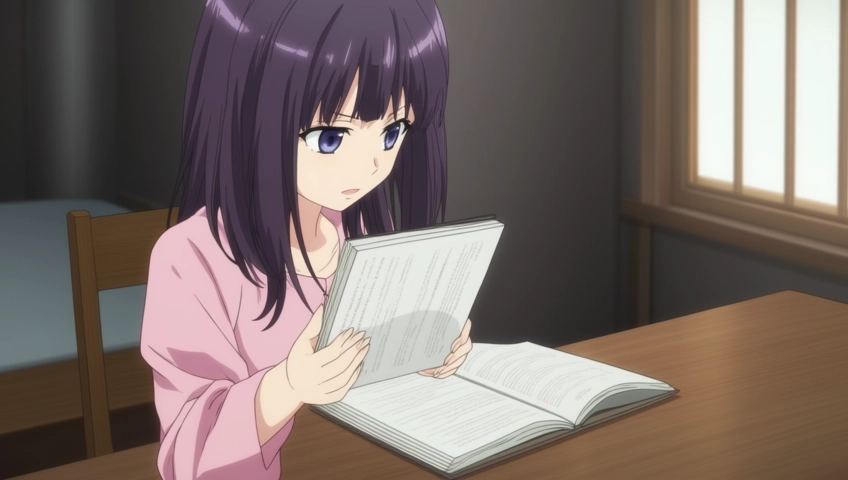} &
\includegraphics[width=0.19\linewidth]{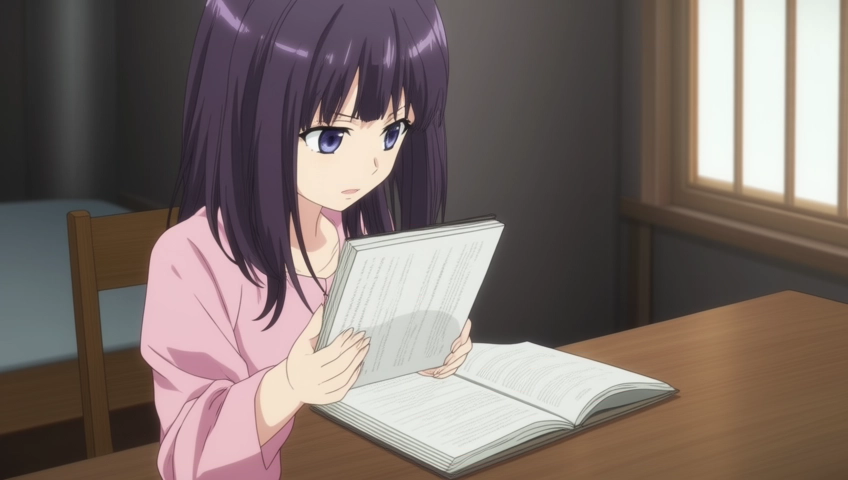} &
\includegraphics[width=0.19\linewidth]{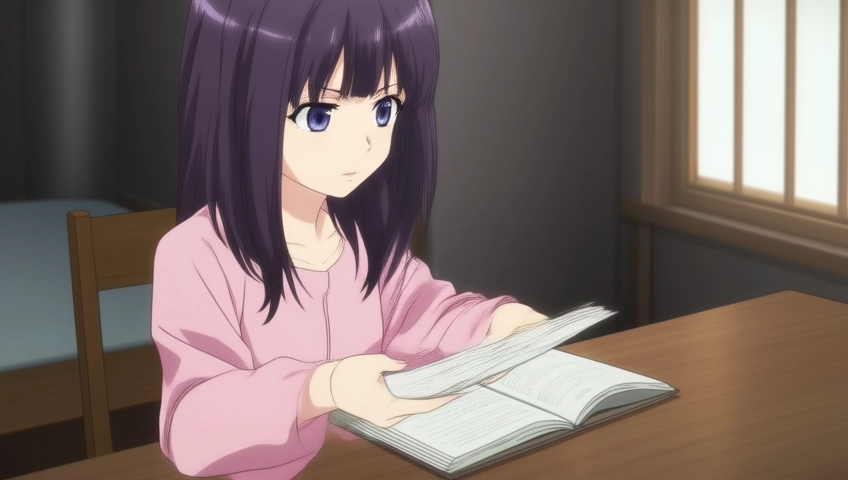} &
\includegraphics[width=0.19\linewidth]{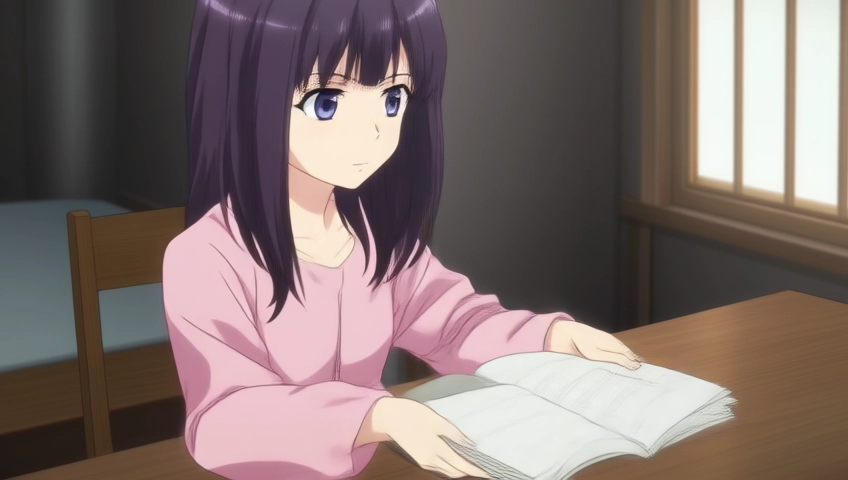} \\
\raisebox{0.0\height}{\rotatebox{90}{\fontsize{6pt}{6pt}\selectfont LTXVideo}} &
\includegraphics[width=0.19\linewidth]{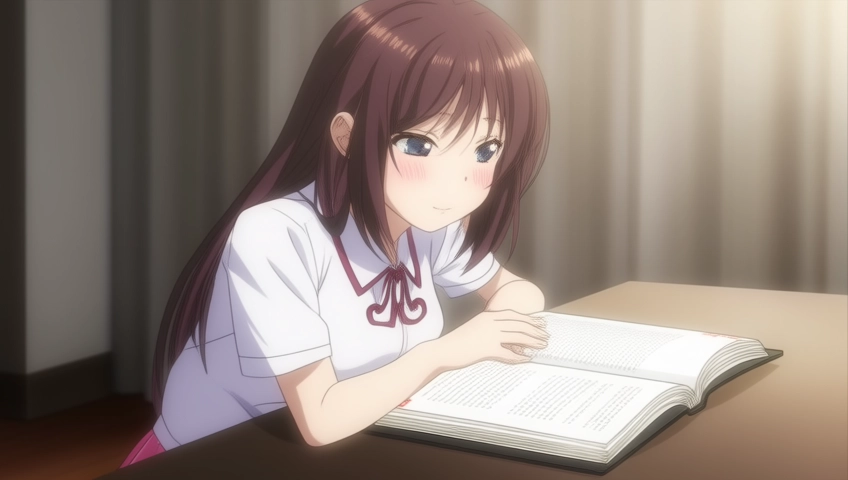} &
\includegraphics[width=0.19\linewidth]{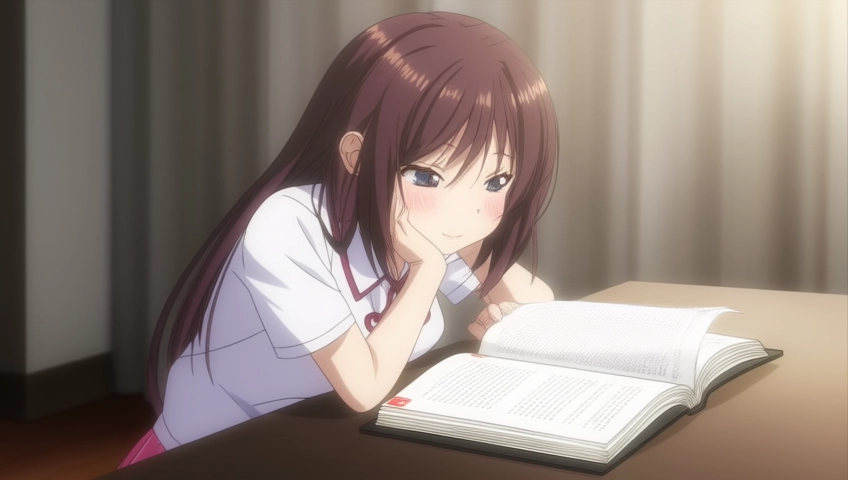} &
\includegraphics[width=0.19\linewidth]{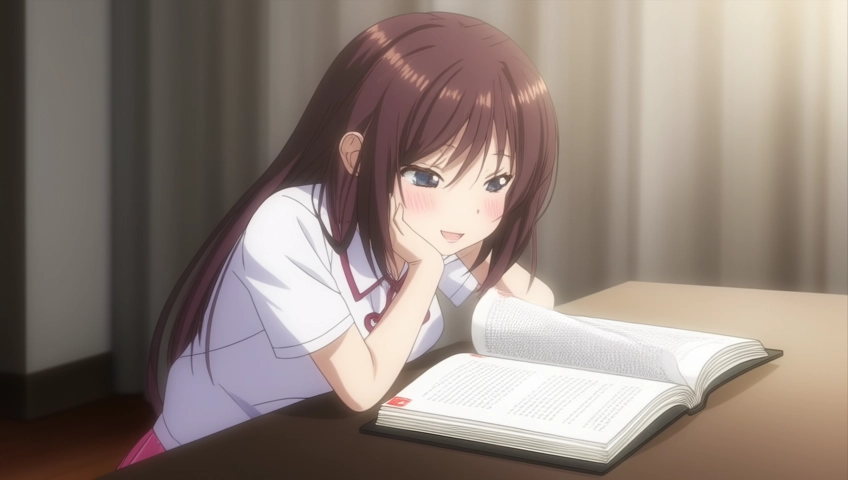} &
\includegraphics[width=0.19\linewidth]{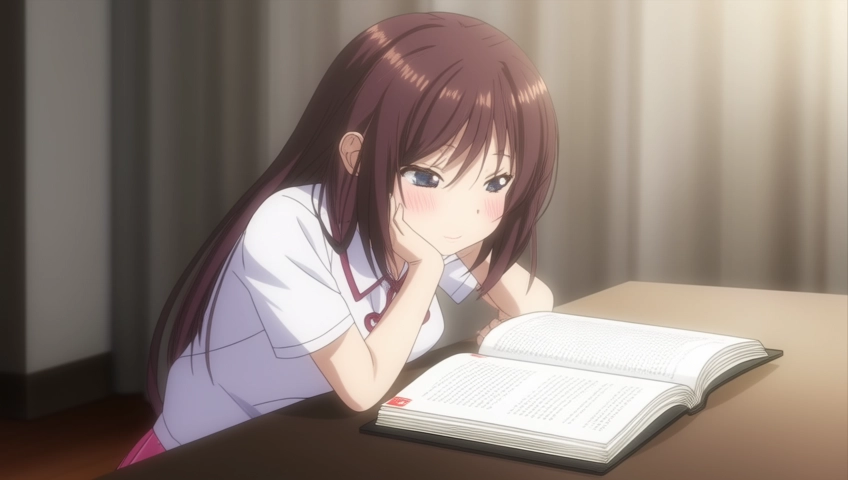} &
\includegraphics[width=0.19\linewidth]{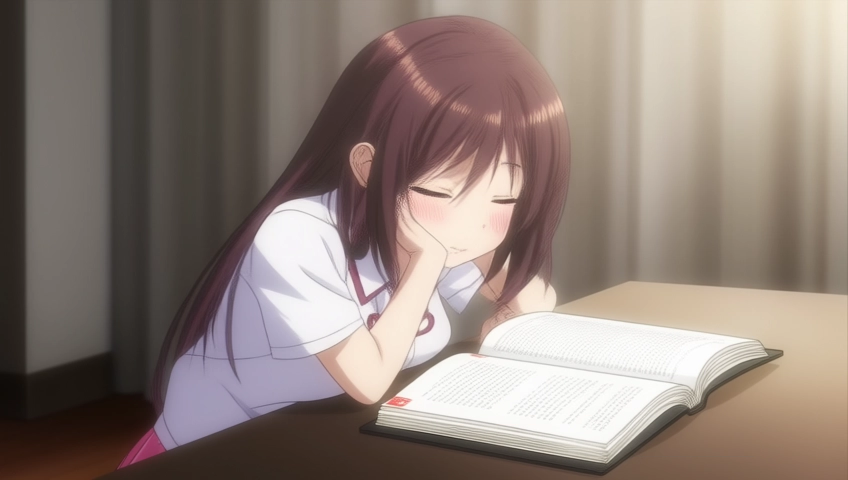} \\
\raisebox{0.2\height}{\rotatebox{90}{\fontsize{6pt}{6pt}\selectfont Wan2.1}} &
\includegraphics[width=0.19\linewidth]{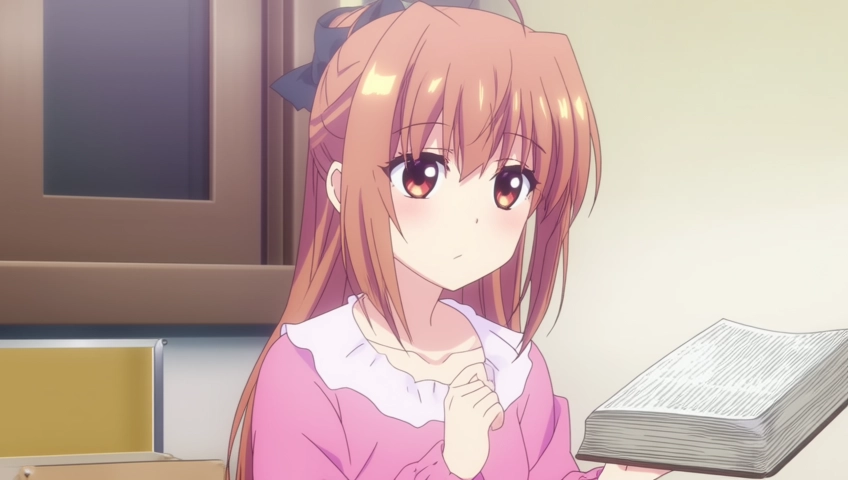} &
\includegraphics[width=0.19\linewidth]{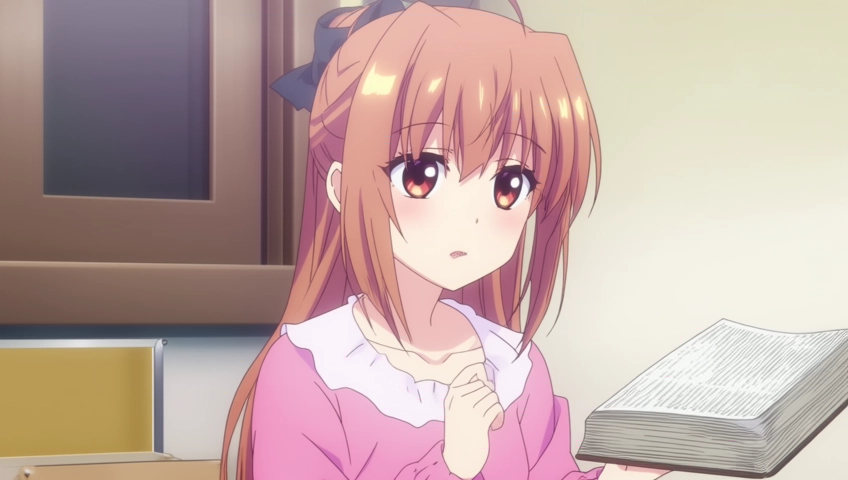} &
\includegraphics[width=0.19\linewidth]{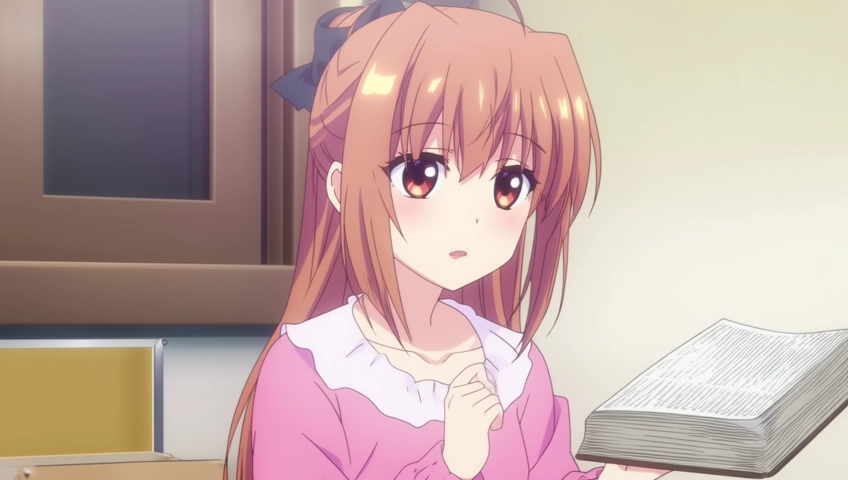} &
\includegraphics[width=0.19\linewidth]{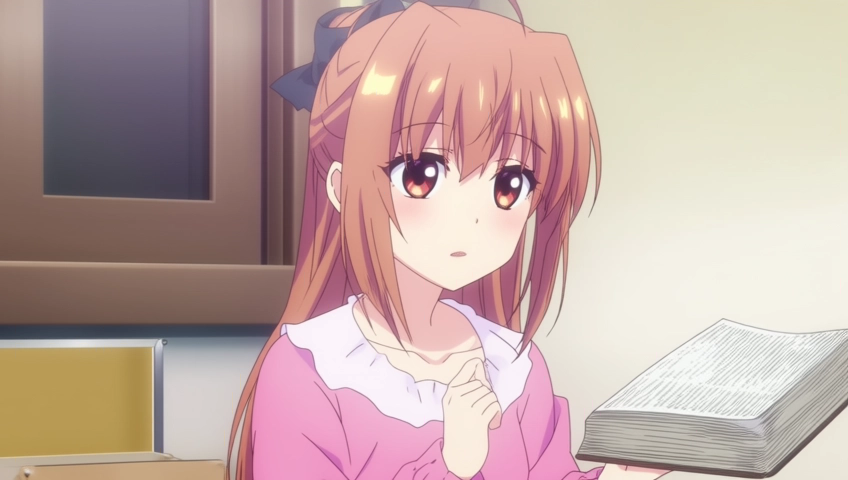} &
\includegraphics[width=0.19\linewidth]{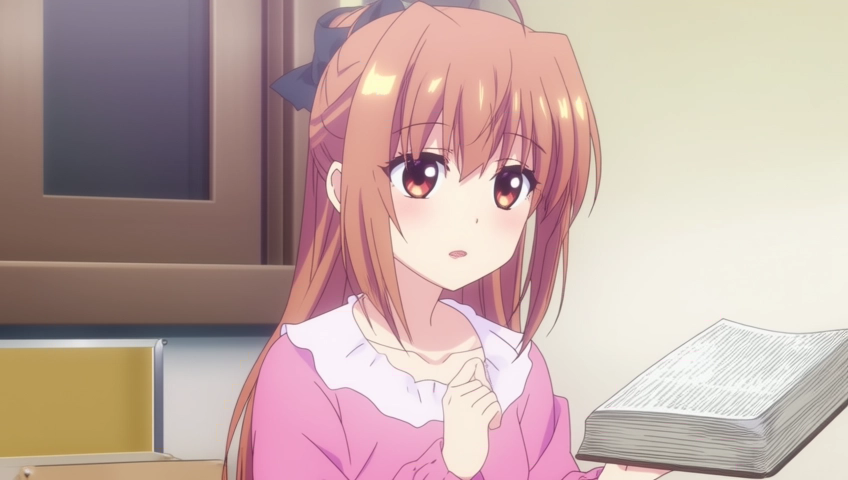} \\
\raisebox{0.2\height}{\rotatebox{90}{\fontsize{6pt}{6pt}\selectfont HYLora}} &
\includegraphics[width=0.19\linewidth]{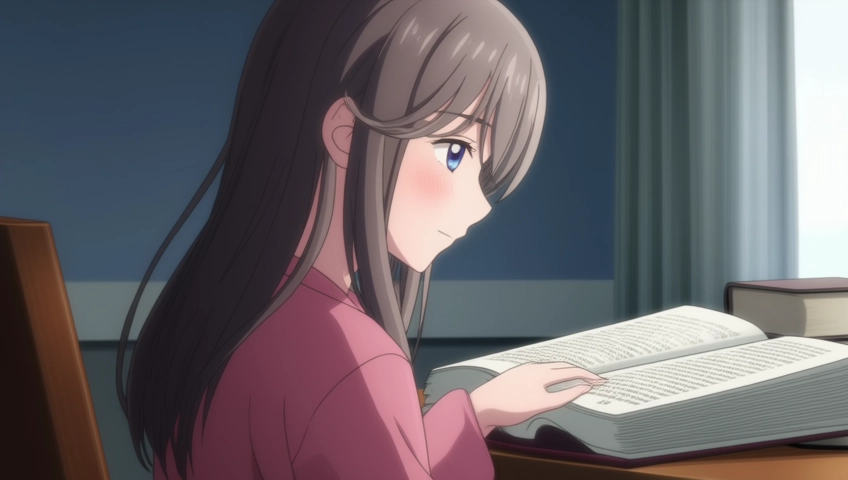} &
\includegraphics[width=0.19\linewidth]{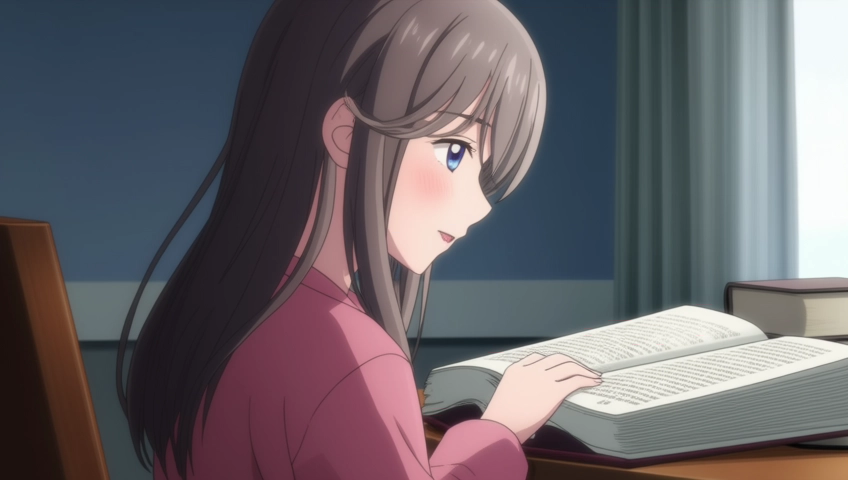} &
\includegraphics[width=0.19\linewidth]{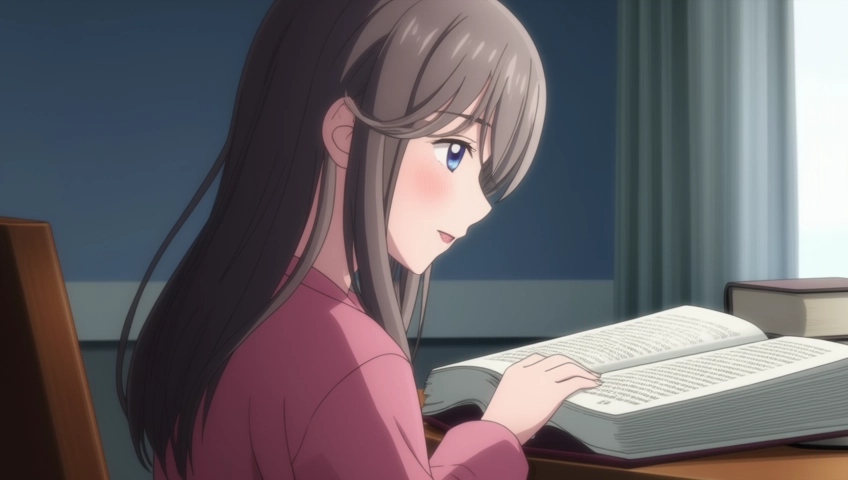} &
\includegraphics[width=0.19\linewidth]{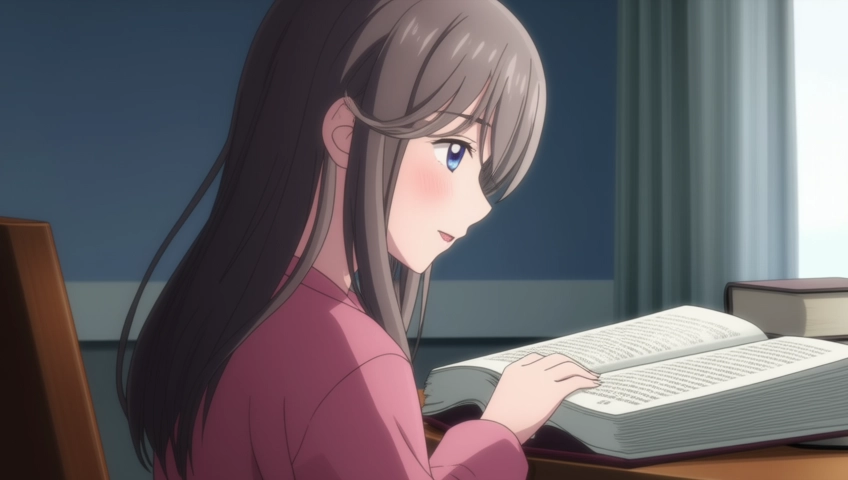} &
\includegraphics[width=0.19\linewidth]{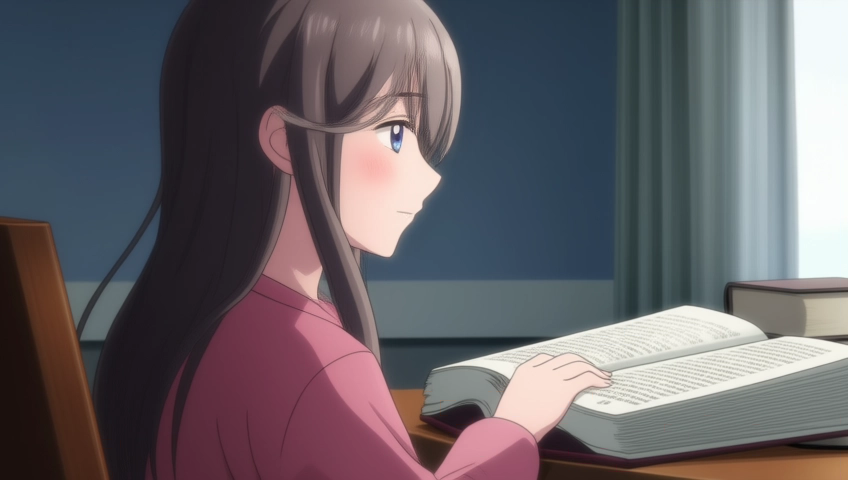} \\
\raisebox{0.5\height}{\rotatebox{90}{\fontsize{6pt}{6pt}\selectfont PTTA}} &
\includegraphics[width=0.19\linewidth]{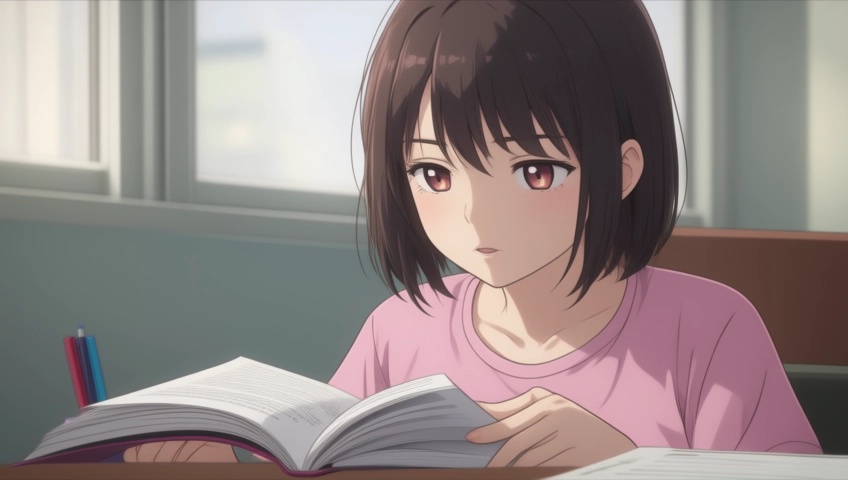} &
\includegraphics[width=0.19\linewidth]{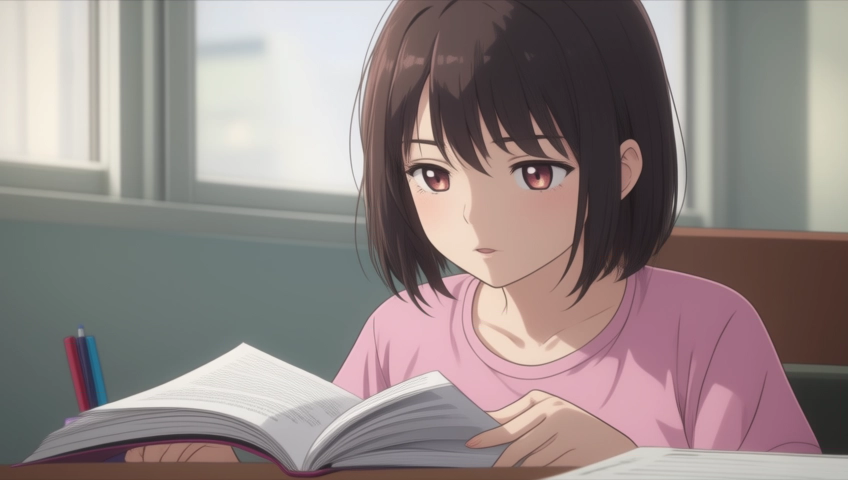} &
\includegraphics[width=0.19\linewidth]{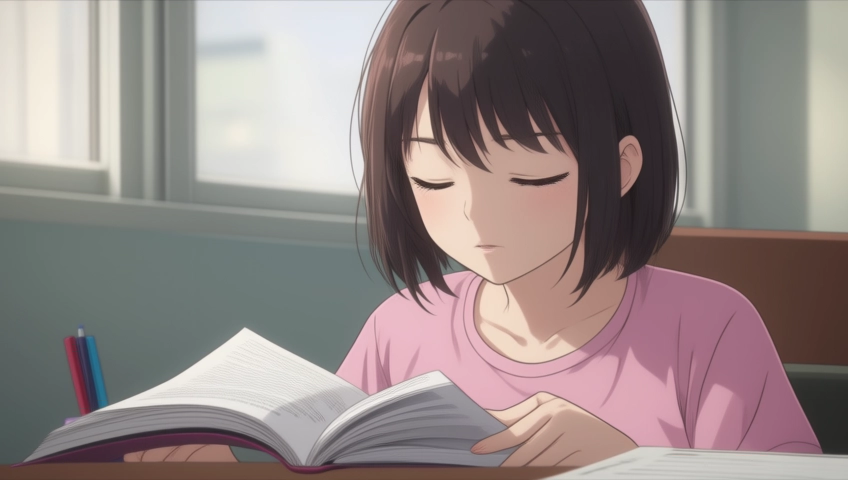} &
\includegraphics[width=0.19\linewidth]{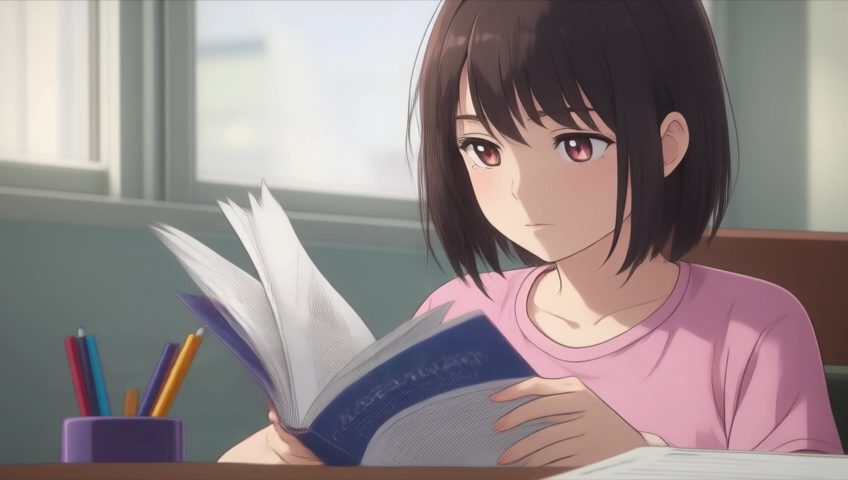} &
\includegraphics[width=0.19\linewidth]{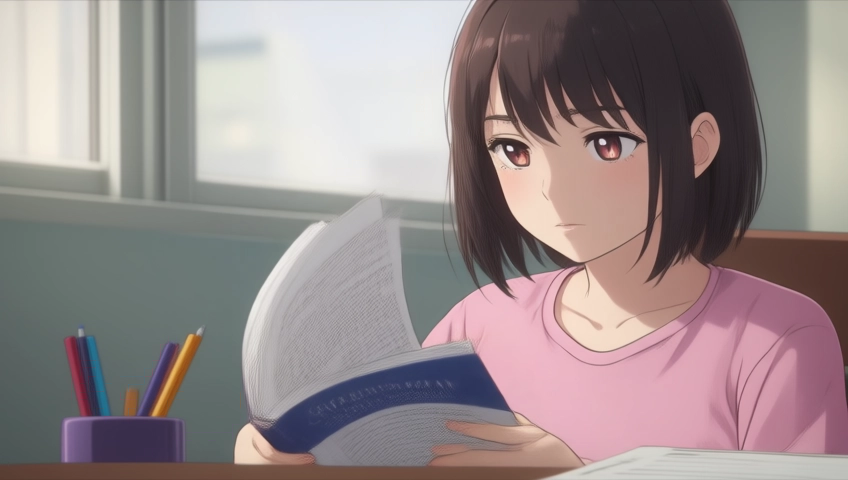} \\
\end{tabular}
\caption{\textbf{Prompt:} anime girl, reading books, desk, gentle action, soft gaze, solo, full body, anime style}
\label{fig:img2}
\end{figure}

\section{Experiment}
\label{sec:experiment}

\subsection{Experimental setup}
\label{ssec:experimentalsetup}

We fine-tuned the HunyuanVideo model using the dataset introduced in Section 3, which contains approximately 12,000 video clips. Each video was resized to a spatial resolution of 800 × 450 pixels, with frame counts ranging from 33 to 97. The model was trained for two epochs. For HydraLoRA, low-rank adapters were inserted only into the linear layers of the Transformer backbone within the diffusion model, with both rank and alpha set to 32. A shared matrix $\mathbf{A}$ and four branch matrices $\mathbf{B}_i$ were used, corresponding to the sub-tasks of male, female, object, and background modeling. During training, only the parameters of the low-rank matrices were updated, while the original model weights remained frozen. Training was conducted on 8 NVIDIA A100 GPUs using the AdamW optimizer, with a learning rate of $2 \times 10^{-5}$, first- and second-order momentum coefficients of 0.9 and 0.99, respectively, and a weight decay of 0.01.

\subsection{Quantitative evaluation}
\label{ssec:Quantitativeevaluation}

For the baseline methods, we select LTXVideo\cite{hacohen2024ltx} Wan2.1\cite{wan2025wan} and HunyuanVideo, three open-source T2V models with strong generative capabilities. To further demonstrate the advantages of HydraLoRA over the conventional LoRA approach, we compare models fine-tuned with both methods, keeping the LoRA parameter settings consistent with a single HydraLoRA branch. Regarding evaluation metrics, since no authoritative benchmark currently exists for text-to-video generation in the animation domain, we construct a dedicated test set consisting of 20 prompts covering diverse animation style elements. For each method, 10 video samples are generated per prompt, resulting in a total of 200 videos. To quantitatively assess the generated results, we adopt VideoScore, which evaluates multiple aspects of video quality, including text-to-video alignment, visual quality, temporal consistency, and dynamic degree. Due to the characteristics of animation, we exclude the factual consistency metric from evaluation.

The quantitative results presented in Table~\ref{tab:benchmark1} demonstrate that our method outperforms baseline approaches across most metrics, including visual quality, dynamic degree, and text alignment. As shown in Figure~\ref{fig:img2}, videos generated by PTTA exhibit superior lighting, detailed rendering, and overall visual fidelity, achieving the highest scores in visual quality. Although the temporal consistency score of PTTA is slightly lower than that of Wan2.1 and HYLora(HunyuanVideo-LoRA), the latter methods produce low-dynamic videos with minimal inter-frame changes, which artificially inflate their temporal consistency scores. In contrast, PTTA generates high-dynamic animations that maintain temporal coherence while remaining highly consistent with the textual descriptions.We also present a qualitative comparison with baseline methods in Figure~\ref{fig:img2}. Animations generated by HYBase(Hun-yuanVideo-Base) exhibit geometric inconsistencies, such as misaligned books, whereas LTXVideo often produces anatomically implausible outputs, like characters with three arms, resulting in suboptimal visual quality. Videos generated by Wan2.1 and HYLora maintain geometric and factual correctness but are relatively static, exhibiting limited motion dynamics.
\setlength{\tabcolsep}{3pt}
\begin{table}[t]
\centering
\captionsetup{position=top}
\caption{Ablation study of the number of $B$ matrices, $N$, in HydraLoRA of PTTA.}
\small
\begin{tabularx}{\columnwidth}{lYYYY}
\hline
\textbf{$B$ matrices nums N} & VSVQ$\uparrow$ & VSTC$\uparrow$ & VSDD$\uparrow$ & VSTVA$\uparrow$ \\
\hline
2 & 2.849 & 2.591 & 2.726 & 2.970 \\
\textbf{4(Default)} & \cellcolor{bestcolor}2.895 & \cellcolor{bestcolor}2.659 & \cellcolor{bestcolor}2.933 & \cellcolor{bestcolor}3.078 \\
8 & 2.773 & 2.480 & 2.719 & 2.696 \\
12 & 2.712 & 2.445 & 2.701 & 2.638 \\
\hline
\end{tabularx}
\label{tab:ablation}
\end{table}

\subsection{Ablation study}
\label{ssec:ablationstudy}
\begin{table}[t]
\centering
\captionsetup{position=top}
\caption{\textbf{Quantitative evaluation.} Metrics are computed via VideoScore~\cite{he2024videoscore}, where higher is better.}
\small
\begin{tabularx}{\columnwidth}{lYYYY}
\hline
\textbf{Methods} & VSVQ$\uparrow$ & VSTC$\uparrow$ & VSDD$\uparrow$ & VSTVA$\uparrow$ \\
\hline
HunyuanVideo(Base) & 2.823 & 2.522 & 2.858 & 2.713 \\
LTXVideo & 2.591 & 1.988 & 2.866 & 2.597 \\
Wan2.1 & 2.847 & 2.673 & 2.714 & 2.891 \\
HunyuanVideo(LoRA) & 2.834 & \cellcolor{bestcolor}2.772 & 2.707 & 2.751 \\
\hline
\textbf{PTTA(ours)} & \cellcolor{bestcolor}2.895 & 2.659 & \cellcolor{bestcolor}2.933 & \cellcolor{bestcolor}3.078 \\
\hline
\end{tabularx}
\label{tab:benchmark1}
\end{table}
The number of $B$ matrices, $N$, in HydraLoRA is a key parameter in our PTTA. An appropriate value for $N$ ensures both training efficiency and model performance. Table~\ref{tab:ablation} presents the experimental results for different values of $N$, with all settings (except $N$) consistent with those in Table~\ref{tab:benchmark1}. Our default value yields the best results. Lower values of $N$ lead to insufficient performance of the fine-tuned model in the new domain, preventing the model from fully exploiting its potential. Higher values of $N$ result in poorer performance due to the excessive number of parameters requiring updates.

\section{CONCLUSION}
\label{sec:conclusion}

We propose PTTA, an animation-style text-to-video (T2V) model capable of generating high-quality animated videos from pure textual input. In addition, we have constructed a high-quality training dataset comprising over 12,000 animation text-video pairs. We expect that both the dataset and the model will contribute to advancing research in this field.

\vfill\pagebreak

\bibliographystyle{IEEEbib}
\bibliography{strings}

\end{document}